\documentclass[sigconf,nonacm]{acmart}

\usepackage{booktabs}
\usepackage{graphicx}
\usepackage{amsmath}

\graphicspath{{figures/}{../figures/}}

\begin{document}

\title{Beyond Model Capacity:
       An Empirical Study of PPO for Portfolio Allocation}

\author{Shun-Wen Lin}
\affiliation{\institution{Academia Sinica}\city{Taipei}\country{Taiwan}}
\email{b11901039@ntu.edu.tw}

\author{Mi-Yen Yeh}
\affiliation{\institution{Academia Sinica}\city{Taipei}\country{Taiwan}}
\email{miyen@iis.sinica.edu.tw}

\author{Wei-Pang Kao}
\affiliation{\institution{Academia Sinica}\city{Taipei}\country{Taiwan}}
\email{r12946017@ntu.edu.tw}

\begin{abstract}
A Mixture-of-Experts agent trained by PPO to allocate Taiwanese equities loses to
buy-and-hold, and this paper asks why. Scored under its own training objective, a typical static
portfolio of fifteen names (the median of $20{,}000$ random draws), bought once and never
traded, outscores every family of networks we trained; and PPO started from a cloned policy that scores well under that
objective lowers the score rather than raising it. Our measurements do not support three
common readings of the shortfall. Held to the policy's own rate of trading, a random
selection of the same number of names is not significantly different from it, while a
momentum rule built from a feature the network already receives, charged the same cost,
beats both. Tuning until the configuration is decisively better on validation leaves the
test-window win rate statistically unchanged. And the identical network, trained by
imitation of that rule, beats the passive bar in most runs with a per-day edge several
times its trading cost, so limited capacity does not account for the shortfall; imitation
is a capacity probe here, not a discovery, since the teacher was chosen knowing it worked
on this window. What the results point to is optimization under the objective. Its
benchmark term, the return of an equal-weight portfolio rebalanced daily, is a baseline
the policy cannot move, changing neither the optimal
policy nor the expected gradient, and on the training years the signal the policy ignores
costs more in turnover than its edge is worth, so departing from the teacher is what the
objective asks for. Two pre-registered screens on the handoff do not prevent the
collapse; one of them, which credits only the names the environment executed instead of
the whole sampled action, ends with a higher sampled objective than its paired baseline in
seven of eight runs and above its starting clone in five, and is the only configuration
here in which this optimiser improves, on average, the quantity it ascends --- a
preliminary result, from eight runs on the training window. The panels are survivor sets,
so absolute returns are optimistic.
\end{abstract}


\maketitle

\section{Introduction}

A reinforcement-learning agent that allocates capital across equities and then
underperforms a passive benchmark admits several readings. It may be undertrained. Its
architecture may lack the capacity to express a profitable policy. Its features may carry
no signal. Or the optimiser may not be improving the objective it was given. These
readings call for different responses, and the literature rarely separates them, because
the comparison that would separate them is not the one usually reported.

This paper separates them on one agent: a Mixture-of-Experts actor-critic --- a policy
network paired with a value head, the \emph{critic}, which predicts future reward and
supplies the baseline PPO scores actions against --- trained with PPO on 103 Taiwanese
equities, holding the top 15 names each day and charged on its turnover. We train 100
models from scratch under an identical protocol and ask four questions.
\textbf{RQ1}: does the trained policy outperform passive and turnover-matched baselines
(Section~\ref{sec:bar})? \textbf{RQ2}: can the architecture represent a profitable policy
(Section~\ref{sec:clone})? \textbf{RQ3}: does PPO preserve or improve a good policy under
its own objective (Section~\ref{sec:rq3})? \textbf{RQ4}: what may make that objective
hard to optimise (Section~\ref{sec:rq4})?

\paragraph{What we observe.} The population loses to buy-and-hold. Held to the policy's
own rate of trading, random selection is not significantly different from it, while a
momentum rule built from a feature the network already receives beats both at the same
charge; tuning the configuration until it is decisively better on validation leaves the
test win rate statistically unchanged; and the identical network, trained by imitation of
that rule, beats the bar in most runs with an edge above its own trading cost. Scored
under the training objective itself, a typical static portfolio that never trades (the
median of $20{,}000$ random draws) outscores every family we trained, and the handoff from a good cloned policy lowers that score. Our
measurements therefore do not support undertraining, absent features or limited capacity
as explanations, and point to optimization under the objective as the open question.

\paragraph{Contributions.} Four measurements, and the instruments that make them.
\begin{enumerate}
  \item \textbf{A turnover-matched evaluation.} Holding trading equal and varying only
        which stocks are picked, we find no significant advantage over random selection,
        and bound it with an interval.
  \item \textbf{Imitation as a capacity probe.} The identical network, trained by
        supervised imitation of a rule known to work on this window, beats the passive
        bar with an edge above its own cost. This is a statement about what the
        architecture can represent, not a discovery about markets.
  \item \textbf{Scoring policies under the training objective itself}, including a
        never-traded baseline. This is what shows that the optimiser is not improving the
        quantity it ascends, separately from any question about what that quantity
        rewards.
  \item \textbf{Zero-update rollouts}, which separate what exploration noise does to a
        policy from what learning does, and show that exploration noise alone lowers a
        good policy's objective before any update.
\end{enumerate}

A companion paper audits the interpretability analyses of the same model population and
reports what does and does not reproduce about its experts; nothing in this paper depends
on it.

\section{Related work}

Deep reinforcement learning for portfolio allocation follows
\citet{jiang2017} in treating allocation as a continuous control problem over a
price tensor, trained on historical returns with a transaction-cost penalty, and PPO
\citep{schulman2017} is the standard optimiser for it. The baseline we find hardest to
beat is the oldest one in the literature: cross-sectional momentum
\citep{jegadeesh1993}, which here is close to a feature the network already receives.

The $1/N$ portfolio is a hard benchmark by construction rather than by accident
\citep{demiguel2009naive}, and trading costs are what decide whether a cross-sectional
signal survives contact with execution \citep{novymarx2016taxonomy}.

Two conventions in that literature matter for what follows. The first is which passive
comparison is used. Buy-and-hold and a periodically rebalanced equal-weight portfolio are
different strategies with different turnover, and they do not generally agree on whether a
policy wins; we therefore report against both throughout. The second is whether the
comparisons are charged at all. A passive or active baseline quoted gross, against a
policy quoted net, measures the cost model rather than the selection. We charge every arm
in this paper identically and report net and gross side by side, which is what
makes the matched-turnover result interpretable rather than merely negative.

Deep-RL portfolio work is commonly benchmarked through shared libraries whose evaluation
harnesses supply the comparison \citep{liu2020finrl}, so the benchmark's definition and
its cost treatment become defaults rather than choices. Our environment is not one of
those libraries but has the same property: its comparison series and its cost convention
arrived with it, which is why Section~\ref{sec:setup} defines both bars before anything is
compared against them.

\section{Setup}
\label{sec:setup}

\paragraph{Model and environment.} The agent is an EIIE-style actor-critic
\citep{jiang2017} whose evaluation head is a Mixture-of-Experts: two shared experts and
four routed experts under top-2 routing. It observes a 20-day window of 70 per-stock
features for each of 103 Taiwanese equities and emits a weight for each. The portfolio
is long-only and fully invested: the 15 largest weights are kept and renormalised to sum
to one before the environment executes them, so no cash is held. Every baseline is fully
invested as well, so differences between arms reflect which stocks are held and how they
are weighted --- the question this paper tests --- and not market timing.
The mixture is incidental here: every result below
is about the objective and the action space, not the head that produces the weights. We
did not train a dense head, so that is an expectation and not a measurement. The experts
themselves are the companion paper's subject.

Training is PPO over $1{,}200$ episodes on an 800-day sampling window;
Appendix~\ref{app:config} gives the hyperparameters. The objective PPO optimises is net
log return in excess of a daily-rebalanced $1/N$ step,
\[
  R_t \;=\; 15\left[\log(1+r^{\text{net}}_t) \;-\; \log(1+r^{1/N}_t)\right],
\]
where $r^{\text{net}}_t$ is the portfolio's return already net of the charge below and
$r^{1/N}_t$ is that day's cross-sectional mean return, i.e.\ an uncharged daily-rebalanced
$1/N$ step. Costs enter only through $r^{\text{net}}_t$: trading is paid for once, and
there is no separate turnover penalty.

The comparison term matters less than it looks. The environment advances one day per step
whatever the agent holds, so $r^{1/N}_t$ depends only on the date: it is a baseline in the
control-variate sense, not a penalty. Over a window it is an action-independent constant,
so $\sum_t R_t$ differs from the net log return by a constant. It therefore ranks policies
identically, leaves the optimal policy unchanged, and contributes nothing to the expected
policy gradient, since $\mathbb{E}[\nabla\log\pi(a\mid s)\,b(s)]=0$ for any $b$ not
depending on the action. It changes variance, not direction. Checkpoint selection inherits
the same property: validation scores each held-out block as return minus that block's
benchmark, a fixed number per block, so ranking checkpoints by excess is ranking them by
return.

\paragraph{The bars, defined once.} Two passive comparisons are available and we report
against both. \emph{Buy-and-hold} is $1/N$ bought on the first test day and held to the
last, weights drifting with returns, worth $+240.1\%$ net over the test window; it has no
turnover after day one, so there is nothing to charge. \emph{Daily-rebalanced $1/N$} is
the standard equal-weight portfolio \citep{demiguel2009naive}, returned to equal weight
every day; charged at the same rate as every active arm here it is worth $+218.1\%$;
uncharged, it is the series the environment ships as its comparison. We take buy-and-hold as primary
because it is the stricter of the two on this window, by $22$ points.

The choice does not carry any conclusion. Of the 100 trained models, $29$ beat
buy-and-hold and $36$ beat daily-rebalanced $1/N$; the turnover-matched random control
beats them $23$ and $47$ times in $200$ draws; the $32$ imitation runs beat them $25$ and
$27$ times. Every population that falls short of a majority against one bar falls short
against the other, and no claim below changes if the bars are swapped. Where the
\emph{uncharged} daily-rebalanced figure of $+219.0\%$ matters it is named explicitly: the
uncharged series is what enters both the step reward above and the validation score, and
in each it enters as a quantity the policy cannot move.

\paragraph{Costs, and what turnover means here.} Turnover is
$T_t=\sum_i|w_{i,t}-w_{i,t-1}|$ over the 103 stocks and cash, measured target to target,
so drift between steps is free. The portfolio is always fully invested, so the cash
position is zero throughout and only the stocks contribute; this sum is not the
$104$-coordinate action of Section~\ref{sec:rq4}, which counts one output per stock plus
an exposure output. The environment deducts $0.0015\,T_t$ from each step's
return: that is $0.3\%$ charged on one-sided turnover --- the fraction of the portfolio
that changes hands, $\tfrac12 T_t$ --- equivalently $0.15\%$ on each leg of a swap, so a sale and the purchase that replaces
it are charged separately. At
the median run's mean daily $T$ of $0.187$ this is $2.81$ basis points. Every active
arm in this paper, baselines included, is charged identically.

\paragraph{Data and splits.} Daily closes for 103 names, 2016-01-04 onward. Training ends
2021-12-31: $1{,}468$ trading days, $1{,}447$ scored steps after the lookback and the
first day's return. Validation is three held-out years scored as the mean excess
over blocks --- 2022 (benchmark $-14.3\%$), 2023 ($+56.0\%$) and 2024 ($+26.3\%$) --- with
checkpoints selected on a trailing mean of the last three validations. The test window is
2025-01-02 to 2026-07-09: 366 trading days, 345 scored steps on the same convention.
Every training-window figure below is scored over those $1{,}447$ steps. All panels are survivor sets; Section~\ref{sec:robust} bounds what
that costs.

\paragraph{Seeds, and what a single run is worth.} Every result is a statistic over a
population of independent runs, never a single run. A run is not reproducible from its
seed alone; five replicates of one seed span $390$ points of test excess
(Appendix~\ref{app:determinism}).

\section{RQ1: Does the policy outperform passive and turnover-matched baselines?}
\label{sec:bar}

The comparison that answers this question holds trading equal and varies only which stocks are
picked, so that what is measured is selection and not the cost model.

\begin{figure*}[t]\centering
\includegraphics[width=\textwidth]{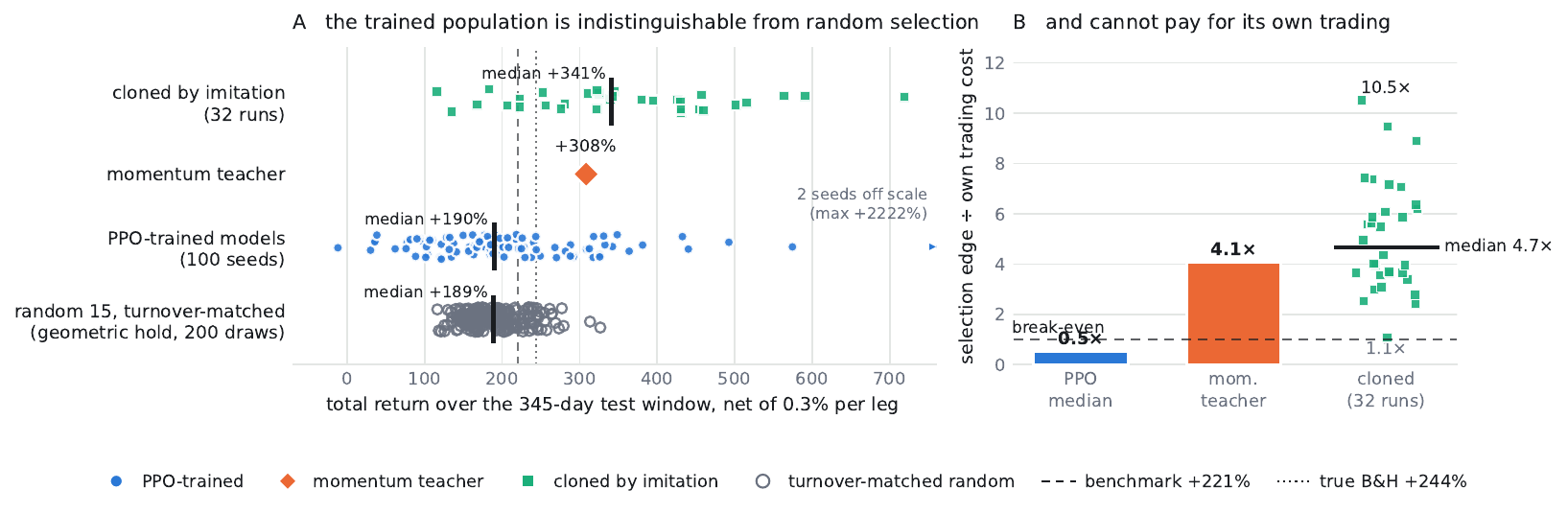}
\caption{\textbf{A.} Net total return over the test window, with every strategy charged
the same trading cost. The 100 PPO-trained models and the 200 random portfolios matched to
their turnover cover the same range, their medians are within a point of each other, and
both sit below both vertical lines. Two trained seeds fall beyond the axis, at $+1128\%$
and $+2222\%$. The cloned row shows every cloning run: $25$ of the
$32$ sit above buy-and-hold and $21$ above the teacher they imitate, median $+341.3\%$,
with the seven that fall short at the left. \textbf{B.} For each strategy, the return per
day its stock picks earn above the universe, divided by what its own trading costs; the
dashed line at $1.0$ is break-even. The PPO models earn $0.53\times$ what their trading
costs. The cloned policies earn $1.1$ to $10.5\times$ theirs, median $4.7\times$, and none
of the $32$ falls below break-even.}
\Description{Two panels. Panel A is a strip plot of total test-window return with four
rows sharing one horizontal axis: 200 turnover-matched random portfolios drawn with a
geometric holding period (median +189\%), 100 PPO-trained models (median +190\%, two beyond the axis at +1128\% and
+2222\%), the single momentum teacher at +308\%, and 32 cloned policies (median +341\%,
spanning +116\% to +719\%). Dashed and dotted vertical lines mark the uncharged
daily-rebalanced 1/N at +219\% and buy-and-hold at +240\%. The random and PPO rows overlap
almost exactly and both sit left of both lines; the cloned row sits largely to the right.
Panel B plots each strategy's per-day selection edge divided by its own trading cost against a
dashed break-even line at 1.0: a bar at 0.5x for the PPO median, 4.1x for the momentum
teacher on this full 103-name panel, and the 32 cloned runs as points spanning 1.1x to 10.5x with a median of 4.7x,
all above break-even.}
\label{fig:returns}
\end{figure*}

\begin{table*}[t]\centering
\caption{Every active strategy pays what the policy pays --- $0.0015\sum_i|\Delta
w_i|$ per step, which is $0.3\%$ of one-sided turnover --- over the test window, holding
15 of 103 names. \emph{Net} is the realised return; \emph{gross} gives the charge back,
so it shows selection without trading. The last two columns count how many members of
each population beat each bar after costs. The turnover-matched random row is the
comparison this experiment was built for: it holds trading equal and varies only which
stocks are picked. The integer-hold row matches turnover more coarsely and is reported
beside it. \emph{Held 11 days} re-selects on the same signal every 11 trading days rather
than daily, which cuts turnover; 11 is the whole-number hold closest to the policy's own
turnover, and the one the integer-hold random control uses.}
\label{tab:costmatched}
\begin{tabular}{lrrrrr}
\toprule
strategy & turnover/day & net & gross & beats B\&H & beats daily 1/N \\
\midrule
\textbf{buy-and-hold (1/N, never rebalanced)} & $0.000$ & $\mathbf{+240.1\%}$ & $+240.6\%$ & --- & --- \\
\textbf{daily-rebalanced 1/N, charged} & --- & $\mathbf{+218.1\%}$ & $+219.0\%$ & --- & --- \\
the same series \emph{uncharged}, used for checkpoint selection & --- & $+219.0\%$ & $+219.0\%$ & --- & --- \\
\midrule
momentum, top 15 by 60-day return, daily & $0.194$ & $\mathbf{+308.4\%}$ & $+351.3\%$ & --- & --- \\
momentum, same 60-day signal, held 11 days & $0.071$ & $+209.3\%$ & $+219.0\%$ & --- & --- \\
minimum volatility, top 15, daily & $0.075$ & $+67.1\%$ & $+73.7\%$ & --- & --- \\
minimum volatility, held 11 days & $0.032$ & $+94.0\%$ & $+97.3\%$ & --- & --- \\
random 15, re-drawn daily & $1.710$ & $+29.9\%$ & $+214.4\%$ & $0/200$ & $0/200$ \\
random 15, held 11 days (integer hold) & $0.160$ & $+192.9\%$ & $+218.4\%$ & $25/200$ & $56/200$ \\
random 15, geometric hold, mean $9.5$ days \textbf{(turnover-matched)} & $0.188$ & $+189.1\%$ & $+217.1\%$ & $23/200$ & $47/200$ \\
\midrule
\textbf{the 100 trained models} (median) & $0.187$ & $+189.7\%$ & $+218.8\%$ & $29/100$ & $36/100$ \\
\bottomrule
\end{tabular}
\end{table*}

Of the 100 models, $29$ beat buy-and-hold and $36$ clear the daily bar. The median model
trails buy-and-hold by $50$ percentage points. The $32$ imitation runs of
Section~\ref{sec:clone} beat the two bars $25$ and $27$ times. Three measurements say what
those counts do and do not show.

\paragraph{Cost explains how far they fall short, not that they fall short.} The median
model trades $0.187$ of its portfolio per day, which costs it $29.4$ points over the
window --- the median of what the 100 models each pay, not the gap between the gross and
net medians of Table~\ref{tab:costmatched}. If we give back the trading costs while keeping exactly the same holdings, the
median excess over buy-and-hold rises from $-50.4\%$ to $-21.3\%$, and the number of
models beating that bar rises from $29$ to $39$. So cost explains much of how far the
models fall short, but not the fact that they fall short: even with no trading cost at
all, fewer than half of them win. The benchmark is charged on the same terms: charged at
the same rate, daily-rebalanced $1/N$ returns $+218.1\%$ against $+219.0\%$ uncharged, a
difference of about one point, so what drags the models down is the cost of their own
trading.

\paragraph{Tuning raises validation performance and leaves the test win rate
statistically unchanged.} We tuned eight hyperparameters, changing
one at a time, and scored each variant by its average excess return over the three
validation years (2022 falling, 2023 bull, 2024 mild), with the rule for choosing the
winner fixed before any variant was scored. The winner was a threefold higher learning
rate, and we retrained it at the full 100 seeds. It is clearly better on the measure it
was selected for: median validation excess $+13.3\%$ against $+6.0\%$ (rank-sum
$p=0.0008$), and $86$ of $100$ seeds beat the uncharged daily-rebalanced $1/N$ across the
three held-out years, against $67$ for the base configuration. On test it beats
buy-and-hold $35$ times against $29$, a difference this design cannot separate from zero
($p=0.36$). Neither $35$ of $100$ nor the $44$ that same configuration would reach with
no trading cost is a majority. A configuration chosen to do better, and which does do better out of sample by
a wide margin, does not win significantly more often on the test window.

The tuning rule has a noise floor of its own, reported in Appendix~\ref{app:tuning}:
re-running the unchanged configuration as a control reproduces most of the margin the
winner was chosen for. The 100-seed comparison above, and not the sweep, is what we rely
on.

\paragraph{At matched turnover we measure no significant selection advantage.} For a fair comparison, the
random portfolios must trade as much as the policy does. Each is 15 names drawn from the
103. Holding one for a fixed eleven days trades too little: $0.160$ per day against the
policy's $0.187$, $14.5\%$ less. Instead we redraw the portfolio with
probability $p$ at each step, which makes the holding period geometric with mean $1/p$.
We choose $p$ using a separate set of random draws, matching only turnover; returns are
never used in choosing it. The 200 reported portfolios are new draws.
At a mean holding period of $9.5$ days these portfolios trade $0.188$ per day against the
policy's $0.187$.

Across 200 such draws the median return is $+189.1\%$ net and $+217.1\%$ gross, against
the models' $+189.7\%$ and $+218.8\%$, and $23$ of the $200$ beat buy-and-hold
against $29$ of the $100$ models (Table~\ref{tab:costmatched}). If we pick one trained
model and one random portfolio at random, the model has the higher gross return with
probability $P=0.520$ ($95\%$ interval $[0.441, 0.600]$) and the higher net return with
probability $0.514$ ($[0.435, 0.594]$), where $0.500$ would mean no advantage; neither
differs significantly from it (rank-sum $p=0.56$ and $p=0.69$). The intervals come from
resampling both the 100 models and the 200 draws (Appendix~\ref{app:pint}), and they are
wide: these data are consistent both with no advantage and with a modest one, so the
measurement bounds the advantage rather than excluding it. The fixed eleven-day
portfolios, undertrading and all, give the same verdict.

A simple rule was profitable on the same window. Holding the 15 stocks with the highest
trailing 60-day return and rebalancing daily, which trades $0.194$ of the portfolio a day
against the policy's $0.187$ and pays the same charge, returns $+308.4\%$: $68$ points
above the bar, and better than $82$ of the $100$ trained seeds. Picking the 15 least
volatile stocks instead returns $+67.1\%$, so not every active rule prospers here. Before
costs, the trained models beat the matched random portfolios by $1.7$ points, which is
well inside the spread of the draws.

\paragraph{The same result measured inside the policy.} The same pattern appears one
level down, before any portfolio is formed. Before the top-15
projection, the policy does rank stocks slightly better than chance: the average daily
rank correlation between its 103 softmax weights and that day's returns (the
\emph{information coefficient}, IC) is $+0.0043$ on
test, and it is positive in $66$ of the $100$ seeds. The size of that number matters more
than its sign. With 103 stocks, a single day's rank correlation varies by about $0.10$ by
chance alone (approximately $1/\sqrt{102}$); averaging over $345$ days, treating days as
independent, reduces that to a standard error of about $0.10/\sqrt{345}\approx0.0054$. A
random ranking scored on the same days has a standard deviation of $0.005$ to $0.007$,
of the same size as this.
And the policy's correlation is $37\%$ of what the same measurement gives the momentum
rule ($+0.0116$). The 15 stocks actually held beat the universe by $+1.50$ basis points a
day on the test window and by $+1.25$, $+3.72$ and $+0.44$ on the 2022, 2023 and 2024
windows, while the median policy's turnover costs it $2.81$ basis points a day on test and
$3.50$, $3.23$ and $2.85$ on those three. On three of the four windows, then --- every one
but 2023 --- the edge is smaller than the cost of acting on it. After the projection the
correlation is not distinguishable from zero at these magnitudes on any of the four
windows ($-0.0010$ on test, then $+0.0015$, $+0.0032$ and $-0.0013$). We report these as magnitudes only. Averaging each model's daily coefficients
into a per-seed mean and testing those means gives $p=0.00018$, but that test is not
valid here: all 100 seeds are scored on the same days and move together, so the effective
sample size is nearer the number of days than the number of seeds.

\paragraph{Takeaway.} \emph{Observed:} at matched turnover the population's advantage over
random selection is $P=0.520$ with a $95\%$ interval of $[0.441, 0.600]$; the names it
holds earn $1.50$ basis points a day over the universe while its trading costs $2.81$;
and a rule built from one of its own features earns more at the same charge.
\emph{Interpreted:} our tests give no support to undertraining or absent features as
explanations, and whatever selection advantage the policy has is smaller than the cost of
acting on it.

\section{RQ2: Can the architecture represent a profitable policy?}
\label{sec:clone}

Section~\ref{sec:bar} reaches
the architecture only indirectly, so we test it directly. We take the identical network
--- same features, same environment, same charge, same 15-name action space --- and train
it by supervised imitation of the momentum rule, whose signal is close to the network's
own feature $\mathrm{close}/\mathrm{SMA}(60)-1$. The loss fits the stock logits to the
teacher's 15 names, which is the only part of the action the top-$k$ projection preserves,
and three DAgger rounds \citep{ross2011dagger} relabel the states the student itself
visits; Appendix~\ref{app:config} gives the rest. Two checkpoints are kept throughout:
the \emph{validation-selected} one, chosen on a trailing mean of the last three
validations, and the \emph{end-of-training} one, taken when cloning stops. We report
both, because validation selection turns out not to help here (Section~\ref{sec:robust}).

\paragraph{It beats the bar, and not by trading less.} Counting every cloning run that
exists, $25$ of $32$ beat buy-and-hold at the validation-selected checkpoint, median
$+341.3\%$, winners from $+252.3\%$ to $+719.0\%$, with seven falling short. At the
\emph{unselected} end-of-training checkpoint the count is $31$ of $32$, median $+420.5\%$,
a single failure at $+165.0\%$. Twenty-one of the 32 beat the teacher they imitate, on net and on
gross alike, while their \emph{teacher overlap} is only $7.4$ to $10.1$ of 15: how many
of a model's 15 holdings are among the teacher's 15 picks on an average day
($15\times15/103\approx2.18$ by chance).
So beating the teacher is common but not universal: four of the 25 winners do not, the weakest
of them returning $+252.3\%$ against the teacher's $+308.4\%$. Risk does not account for
the result but it does spread widely: annualised Sharpe runs from $1.22$ to $3.31$ against
the teacher's $2.53$, and maximum drawdown from $27\%$ to $39\%$ against its $35\%$. Per-day selection edge runs from
$1.1$ to $10.5$ times each run's own cost, median $4.7\times$, with none of the 32 below
break-even, against $0.53\times$ for the PPO population (Figure~\ref{fig:returns}B).

\paragraph{They are not simply the teacher.} Under a one-day execution delay that drops
the momentum rule $25.4$ points \emph{below} buy-and-hold, five of the six runs
re-evaluated under that delay still beat it, at $+267.7\%$ to $+556.5\%$; the sixth falls
to $+92.0\%$. Whatever those five learned is more robust to execution timing than the rule
they copied. The delayed arm covers six of the 32 runs, not all of them.

Results are counted per run, not per seed (Appendix~\ref{app:determinism}). The rate also
falls as the population widens --- $16$ of the first twenty,
nine of the last twelve --- and we report the pooled figure and both parts rather than
the first.

\paragraph{Takeaway.} \emph{Observed:} $25$ of $32$ imitation runs beat buy-and-hold at
the validation-selected checkpoint and $31$ of $32$ at the end of training, every one of
the $32$ with a per-day edge above its own trading cost, and five of the six re-evaluated
under a one-day delay survive a delay that puts their teacher below the bar.
\emph{Interpreted:} the architecture can represent a policy that beats the bar when
trained by imitation of a known rule. That is a statement about representational
capacity, not a discovery: the teacher was chosen because we had already measured that
rule beating this benchmark on this window, so cloning it and reporting that the clone
works is not evidence of out-of-sample discovery.

\section{RQ3: Does PPO preserve or improve a good policy under its own objective?}
\label{sec:rq3}

Section~\ref{sec:clone} leaves a policy that scores well under the training objective.
This section hands it to the optimiser and measures the result under the objective PPO is
itself ascending.

\paragraph{PPO improves the ranking and does not reliably keep the portfolio.} We
give the cloned weights back to the normal PPO training loop for 600 episodes. In the
three runs without a critic warm-up, PPO improves the network's ranking of all 103 stocks
before the top-15 cut --- the information coefficient of Section~\ref{sec:bar}, and the
output its update is computed on. But teacher overlap falls: the 15 stocks it actually
holds drift away from the teacher's. This is consistent with PPO improving the output it is trained
on rather than the 15 stocks that are traded. Across all eight handoffs, teacher overlap at the
end of training is $7.16$, $2.37$, $3.15$, $2.53$, $1.88$, $1.77$, $2.33$ and $1.44$ of
15, against $2.18$ by chance: seven of the eight end close to chance, and we cannot
explain the run that keeps $7.16$. The median return falls from $+387.4\%$
at the cloned starting point to $+199.3\%$ with the validation-selected checkpoint and
$+141.8\%$ at the end of training; $7$ of $8$ runs beat buy-and-hold before the handoff
and $3$ of $8$ after, under either checkpoint rule.

\paragraph{A warmed critic does not prevent the collapse.} The imitation loss never touches the critic,
so the handoff begins with a trained actor and an untrained value head, and the collapse
has two candidate explanations. Five further runs repeat the handoff with the value head
first regressed on the cloned policy's own rollouts and the actor frozen. The warm-up
does not help: teacher overlap ends at $2.53$, $1.88$, $1.77$, $2.33$ and $1.44$ of 15,
against $2.18$ by chance; the pre-projection rank IC ends negative in
four of the five, and the return outcome splits as it did without warming. The
pre-registration allowed the warm-up to delay rather than prevent the collapse; that is
not supported either, since all five warmed runs end at chance. These five runs give no
evidence that critic-initialisation shock is what removes the cloned policy. Three limits
on that reading. The
warm-up did not fully train the critic: prediction error fell $18$--$36\%$ in four runs
and rose in the fifth. Training is not reproducible from a seed, so warmed and unwarmed
runs are separate draws, not matched pairs. And only the first two of the five were
pre-registered.

\paragraph{What the objective rewards, and what the benchmark term does.}
With critic initialisation unsupported as an explanation, it is worth being exact about
what the reward says. $R_t$ of Section~\ref{sec:setup} does contain a comparison series, but as a baseline
the policy cannot move: within a window it shifts every policy's score by the same
constant, so it changes neither the optimal policy nor the expected gradient. Ranking
policies by $R$ on a fixed window is ranking them by net return. The question is therefore
not whether momentum beats a benchmark on the training window but whether it earns more
than the alternatives there, and it does not.

Rolled through the \emph{training} period, 2016 to 2021, the rule returns $+531.8\%$
before costs and $+298.9\%$ after them, trading $0.212$ of the portfolio a day over the
$1{,}447$ scored steps. It pays away roughly $233$ points to its own trading. Against the
daily-rebalanced $1/N$ series that leaves it $-50.0\%$ after costs and $+182.9\%$ before
them. The $-50.0$ is \emph{momentum's own cost}, not a handicap the benchmark imposes. A
rule that spends more on turnover than its gross edge is worth loses to holding under any
net-return objective, and charging the benchmark would change nothing, because it would
shift every policy's score by the same constant and leave the ranking unchanged. Table~\ref{tab:trainreward} scores the arms under the objective
as implemented.

\begin{table}[t]\centering\small
\caption{Training-window score under the reward as implemented,
$R=15\sum_t[\log(1+r^{\text{net}}_t)-\log(1+r^{1/N}_t)]$, over the $1{,}447$ scored steps.
Medians within each family. $R=0$ is the comparison series itself. The population row is
the validation-selected checkpoint; from end-of-training weights it scores $-9.75$. The
cloned row is all $32$ clones at their validation-selected checkpoint, and the handoff row
the eight of them that were handed back to PPO. A
static 15-name portfolio, drawn once and never traded, outscores every trained family.}
\label{tab:trainreward}
\begin{tabular}{lrr}
\toprule
family & train return & median $R$ \\
\midrule
static random 15, never traded ($n{=}20{,}000$) & $+377.3\%$ & $\mathbf{+0.92}$ \\
cloned policies, selected (all $32$) & $+367.0\%$ & $+0.59$ \\
\emph{the comparison series itself} & $+348.9\%$ & $0.00$ \\
momentum teacher & $+298.9\%$ & $-1.77$ \\
handoff from the clones ($n{=}8$) & $+157.5\%$ & $-8.40$ \\
the PPO population, selected ($n{=}100$) & $+116.0\%$ & $\mathbf{-10.98}$ \\
\bottomrule
\end{tabular}
\end{table}

This separates two things that are easy to run together. The first is objective design:
on the training years momentum loses to the comparison series after costs, so a policy
that learns to ignore it is doing what its objective asks, and departing from the teacher
is not a failure. The second is optimization: the trained models score far below a
portfolio that simply buys 15 random stocks and never trades. The median gap between the
PPO population and the static random draw is $11.90$ units of $R$. Splitting it into a
trading term and a selection term puts $6.40$ units on trading and $4.61$ on selection,
about $54\%$ and $39\%$ of the gap; the two do not add to the whole because each is a
median taken separately. So more than half of what PPO gives up is paid for trading that
does not earn its cost back.

\paragraph{Takeaway.} \emph{Observed:} under its own training objective the PPO population
scores $-10.98$ against $+0.92$ for a never-traded static portfolio; the eight handoffs
end at a median $R$ of $-8.40$ under that objective, against $+0.89$ for the same eight
clones before the handoff (Table~\ref{tab:trainreward} gives $+0.59$ for all $32$); and
$7$ of $8$ runs beat buy-and-hold before the handoff against $3$ of $8$ after it, with
teacher overlap falling to chance in seven of the eight. \emph{Interpreted:} PPO lowers the objective it
optimizes, starting from a policy that scores well under it. That is separate from what
the objective rewards: leaving the teacher is what the objective asks for on this window,
but scoring below a portfolio that never trades is not.

\section{RQ4: What may make this objective hard to optimise?}
\label{sec:rq4}

This section is diagnostic and its
mechanisms are hypotheses unless stated otherwise: we measure two properties of the
setup, then test the repair each one predicts.

\paragraph{Exploration noise lowers the objective before any update.} To show this we roll the eight cloned networks, and
ten of the PPO population, over the training years twice with \emph{no update applied}:
once without exploration noise, and once with the noise PPO adds during training, 20
rollouts each. The objective falls and turnover rises in
all 18. The stocks held barely change: median teacher overlap falls only from $13.68$ to
$13.18$ of 15 (at most $0.56$). What changes is trading. Daily turnover rises $2.3\times$
for the clones, from $0.215$ to $0.499$ of the portfolio per day, and $3.4\times$ for the
PPO models. The likely mechanism, which we do not measure directly, is that the noise
jitters weights across the top-15 cutoff, so the executed set churns from day to day
while its average composition survives. Returns are net of
cost, so that churn alone lowers the clones' median objective from $R=+0.89$ to
$R=-8.59$, about $10$ units --- a median over the eight handoff clones, each scored on a
deterministic rollout of the training window and then on the same rollout with noise. The noise scale responsible was never chosen: cloning's loss
has no term in the exploration spread, so every handoff starts at the same default of
$\sigma=0.368$, and $1{,}200$ episodes of PPO leave it within $0.05$ of that. Under exploration noise,
then, the cloned policy's objective is already far below its deterministic value before
any update is applied.

\paragraph{The optimiser credits an action the environment mostly discards.} The
projection keeps the 15 largest stock weights, renormalises and zeroes cash. The policy
samples a raw $104$-dimensional action --- 103 stock-selection logits and one exposure
logit, which the projection discards --- and PPO's importance ratio is computed over all
$104$. So $89$ of the $104$ coordinates carry gradient with no effect on any price: the
exposure logit cancels in the renormalisation, and the other $88$ count only by being
below the cut, while the 15 that executed are credited only through the pre-projection
density. We state as a hypothesis, not a measurement, that this is why the measured
ranking survives only before the projection.

\subsection{Two handoff screens}
\label{sec:screens}

The two mechanisms above --- an unchosen exploration scale and credit assigned through an
action the environment mostly discards --- each predict a repair, and we pre-registered
both as screens, fixing the decision rule and the predictions in advance. \textbf{(B1)}
drops the exploration scale to $\sigma=0.100$ after loading the clone. \textbf{(A)}
changes how PPO measures the size of each update. PPO compares how likely the stored
action is under the new policy and under the old one --- the importance ratio --- and
clips updates that change it too much. Normally that probability covers the whole raw
action, all $104$ numbers the network outputs. Screen A instead scores only the 15 stocks
actually traded, in the order they were picked: the chance of drawing the first, then the
second from those still available, and so on: a Plackett--Luce probability. Outputs that
never trade no longer affect the update. Nothing else
changes in either arm; both run the same eight cloned checkpoints for 600 episodes, paired
run for run with the baseline. That baseline is a fresh set of eight handoff runs, started
from the same eight clones as Section~\ref{sec:rq3} and logged alongside the screens;
because training is not reproducible from a seed (Appendix~\ref{app:determinism}) its
numbers are not those of Section~\ref{sec:rq3}, and only the paired comparison within this
experiment is read. The registered rule called collapse prevented for a run if
final teacher overlap reached $7$ of 15 \emph{and} final $R$ stayed within $2$ units of
that run's own clone, and called a screen successful at $6$ of 8 runs.

\textbf{Neither screen prevents collapse: $0$ of $8$ in both arms, as in the baseline's
$0$ of $8$.} Lowering the exploration scale alone did not prevent the collapse, and B1 is
not directionally better than the baseline --- mean final teacher overlap $2.70$ against
$3.23$ of 15 --- while cutting $\sigma$ by $3.7\times$ moved sampled turnover over the
first ten episodes only from $0.478$ to $0.351$ and raised the fraction of updates
meeting the clip boundary from $4.8\%$ to $41.8\%$, in 8 of 8.

\paragraph{A secondary observation the pre-registration did not anticipate.} Screen A is
the only arm in which PPO improves its own objective. On the training window, over $8$
runs started from clones, the sampled objective --- the quantity the gradient sees ---
rises above the paired baseline's in $7$ of $8$ runs ($p=0.0156$, a two-sided exact
Wilcoxon signed-rank test over the eight paired runs, as for every paired $p$ in this
section), and the deterministic
training-window score ends at a mean of $R=+5.09$, against a mean of $+1.40$ for the
eight clones the screens start from, measured at their first episode (the median of the
same eight is the $+0.89$ of Section~\ref{sec:rq3}), and a mean of $-9.85$ for the
baseline handoff. Five of the eight screen-A runs end above their own starting clone
(Table~\ref{tab:screenruns}). Quieter exploration is not what does this: screen A trades the least of
any arm while its exploration scale \emph{rises}. We read that as the executed-set score
rewarding wider logit gaps at the top-15 cutoff, which would make a selection survive
being sampled; that reading is an interpretation, not a measurement. Its importance
ratios are also far less spread out (Appendix~\ref{app:screens}). But it drops the
teacher outright --- mean final teacher overlap $0.38$ of 15, below the paired baseline in 8 of
8 ($p=0.0078$, the exact two-sided floor at $n=8$) --- and it buys no selection advantage,
the pre-projection IC ending at $+0.0002$. We interpret the departure from the teacher as
what the objective asks for rather than as a failure: on the training years the teacher
scores below the series the reward measures against, so a policy that maximises the
objective should leave it. The registered rule required the teacher's names and the
objective to improve together, so it has no cell for this. These are preliminary
training-window results from $8$ runs, not evidence about test returns.
Appendix~\ref{app:screens} has the per-run table, the remaining readouts and the one
prediction that cannot be scored.

\paragraph{Takeaway.} \emph{Observed:} with no update applied, exploration noise raises
the clones' daily turnover $2.3\times$ and lowers their median objective by about $10$
units while the stocks held barely change; lowering the exploration scale does not prevent
the collapse ($0$ of $8$); and crediting only the executed names ends with a higher sampled
objective than the paired baseline in $7$ of $8$ runs on the training window, and above
its own starting clone in $5$ of $8$. \emph{Interpreted, as
hypotheses:} churn at the top-15 cutoff under noise, and credit assigned through
coordinates the environment discards, are the two candidate reasons this objective is
hard to ascend; the second is the one an intervention moved, at $n=8$ and on the training
window only.

\section{Robustness}
\label{sec:robust}

\paragraph{The momentum baseline is not an artefact of one lookback or one window.}
We run the rule over a grid: seven lookbacks from 5 to 250 days, two holding periods and
all four test windows. Each combination is one \emph{cell}, charged exactly as
every other arm in this paper is (Table~\ref{tab:momgrid}). The rule
beats buy-and-hold in $36$ of $56$ TAIEX cells, and in a majority of cells in three of the
four windows. At every lookback tested, rebalancing daily, it also beats at least $74$ of the $100$
trained seeds on the test window. Its one failure is the falling window of 2022, where a
cross-sectional trend rule loses at all seven lookbacks before costs as well as after ---
the familiar momentum crash. The rule's own level depends strongly on the lookback: on the
test window its excess over buy-and-hold runs from $+7$ points at a 120-day lookback
to $+424$ points at a 20-day one. So $+308.4\%$ is one point on that surface and not a
tuned competitor.

\paragraph{Checkpoint selection, and all four test windows.}
Table~\ref{tab:fourwin} scores the 100 trained models on each window from their
\emph{end-of-training} checkpoint: the weights as training left them, with no validation
score used to choose when to stop. 2022 is the one window where the population is the
stronger of the two arms, its median excess beating momentum's there by $7.4$ points and
nowhere else, and it is also the one window where momentum loses to the bar. The
population still loses to buy-and-hold on 2022 itself. From end-of-training weights the
population's median excess is $-96.3$ points, with $18$ of $100$ above buy-and-hold; the
validation-selected figures of Section~\ref{sec:bar} are $-50.4$ points and $29$ of
$100$, and under the training objective the same two checkpoints score $-9.75$ and
$-10.98$ (Table~\ref{tab:trainreward}). So validation selection is worth about $46$
points of test return to the PPO population, while costing it about a unit of the
objective it was trained on.

For the cloned runs, validation selection does not help. Among the first six, the run with
the \emph{highest} validation score is the only one that loses on test and the one with
the lowest is the best (rank correlation $\rho=-0.77$, $p=0.072$, over those six runs);
across the whole arm the end-of-training checkpoint beats buy-and-hold more often than the
validation-selected one, by $19$ against $16$ among the first $20$ runs and $31$ against
$25$ over all $32$. That is why both checkpoints are reported throughout.

\paragraph{The charge is lighter than a real one, and the ordering survives a real one.}
The environment charges $0.15\%$ of every leg traded (Section~\ref{sec:setup}). Real
Taiwanese retail trading costs more than that. Commission is $0.1425\%$ a side, often discounted to
$0.05$--$0.10\%$, and a $0.3\%$ transaction tax is charged on sales. Those components
together come to roughly $0.20$--$0.29\%$ a leg. We recost every arm at both ends of that
range from its gross return and mean turnover (Table~\ref{tab:costsens}). At the
harsher rate the ordering of the arms is unchanged: momentum returns $+271.7\%$ against
buy-and-hold's $+239.6\%$, the trained population's median falls to $+161.6\%$ with $22$
of $100$ still above buy-and-hold, and $22$ of the $32$ cloned runs still clear it.
Buy-and-hold barely moves, because it pays for one purchase and never trades again. So a
heavier charge narrows every active margin without reordering anything.

\paragraph{A second market.} We run the same grid on 98 S\&P~100 names. There the momentum rule beats
buy-and-hold gross in $39$ of $56$ cells but net in only $24$. The charge is why: for
names of that size $0.15\%$ a leg is about three times a realistic cost, because
commissions are far lower and there is no sell-side transaction tax, and at $0.05\%$ per
leg it beats net in $37$ cells. We report the market at the Taiwanese rate because that is
what makes it like-for-like against our own runs. No S\&P~100 net return quoted here is a
realistic net return for that market.

\paragraph{Both panels are survivor sets, and no within-universe control can see it.}
A battery of controls that all hold one thing fixed cannot detect a bias in that thing.
Every one of the 103 Taiwanese names has a valid price on every one of $2{,}561$ days, and
every one of the 98 S\&P~100 names on all $3{,}147$: both are panels of survivors. So
every absolute return here, buy-and-hold's included, is inflated,
and no control that stays inside the universe can see that, because they all share the
universe. We detect this by counting missing prices: there are none.

What survivorship hands every strategy is the knowledge that these names survived. One way to take
part of that back is to delete the best performers over the full sample, and doing so
\emph{widens} momentum's edge over the uncharged daily-rebalanced
$1/N$ series: from $+89.3$ points to $+102.9$, $+111.8$ and $+126.8$ points as the best 5,
10 and 20 names are removed. But that test cannot put back what a survivor panel removes,
which is names that fell and stopped trading. So Appendix~\ref{app:delist} injects such
names instead, delisting $k$ names on randomly chosen days at a terminal loss of $X\%$ and
making them untradeable from then on, over a sweep of both. Momentum's median margin over
buy-and-hold stays positive in every cell of that sweep, and the trained population reaches a majority in none of them. Buy-and-hold itself
falls much further than the margin does, because a passive portfolio holding all 103 names
cannot rotate out of a name that is failing. Appendix~\ref{app:pretest} re-runs the
comparisons on the names that were on the universe list before the test window opened.

\begin{table}[t]\centering\footnotesize
\caption{The population on all four test windows, from each run's end-of-training
checkpoint, so every window after 2021 is out of sample for it. \emph{Median} excess in points over each window's own buy-and-hold;
$n$ is the count of the 100 runs above that bar. The random column is the integer 11-day
hold, the only random arm available on all four windows; on the test window the
turnover-matched geometric arm gives $-51.0$ rather than $-47.2$.}
\label{tab:fourwin}
\begin{tabular}{lrrrrr}
\toprule
window & B\&H & pop. & $n$ & mom. & rand. \\
\midrule
2022 \emph{fall} & $-13.0\%$ & $-3.9$ & $42$ & $-11.3$ & $-5.8$ \\
2023 \emph{bull} & $+56.4\%$ & $-20.6$ & $21$ & $\mathbf{+52.1}$ & $-10.7$ \\
2024 \emph{mild} & $+32.9\%$ & $-27.1$ & $14$ & $\mathbf{+24.0}$ & $-13.1$ \\
test & $+240.1\%$ & $-96.3$ & $18$ & $\mathbf{+68.3}$ & $-47.2$ \\
\bottomrule
\end{tabular}
\end{table}

\section{Discussion, limitations and open questions}

\paragraph{What is observed.} The population loses to buy-and-hold. At matched turnover
its advantage over random selection is $P=0.520$, with an interval wide enough to include
both no advantage and a modest one, while the names it holds earn less per day than its
own trading costs. A configuration tuned until it is decisively better on validation does
not win significantly more often on test. The identical network trained by imitation beats the bar in $25$ of
$32$ runs with an edge above its own cost in all $32$. Under the training objective, a
static portfolio that never trades outscores every trained family, and the handoff from a
cloned policy lowers that objective rather than raising it. A warmed critic does not
prevent that. Neither of two pre-registered screens prevents it either; the one that
credits only the names the environment executed ends with a higher sampled objective than
its paired baseline in $7$ of $8$ runs and above its starting clone in $5$, on the training
window and from eight runs.

\paragraph{What can reasonably be concluded.} Our tests give no support to
undertraining, absent features or limited capacity as explanations, within what each test
covers: one tuning sweep, one feature-based rule, one imitation teacher. The
objective's comparison term is a baseline the policy cannot move, so the shortfall is not
an artefact of an unfair benchmark, and the signal the policy ignores is one the objective
gives it no reason to hold on the training window. What the results point to is
optimization under the training objective: the optimiser does not improve the quantity it
ascends, from a starting point that scores well under it. Objective design and
optimization failure are separate, and the evidence here concerns the second.

\paragraph{What remains open.} Why PPO lowers this objective is not established; this
paper narrows the candidates and tests two of them. Training from scratch under
executed-action credit assignment, with its own pre-registration, is the experiment we
have not run, and the screen that motivates it is $n=8$ on a training window. Whether a
dense head behaves differently is untested: we did not train one. Whether any of this
survives a point-in-time universe is unknown, and only such a universe would settle the
survivorship question. Appendix~\ref{app:winners} reports one further property of this
population --- the models that beat the bar hold more alike than chance --- which we
leave open rather than read as a selection advantage, because untrained networks show it too.

\paragraph{Limitations.} Everything here is simulated. Returns are backtested on historical closes with no
slippage, market impact or partial fills, and charged a flat rate regardless of order
size; one cost rate is applied to two markets, which keeps comparisons like-for-like
\emph{within} each market only.
Every strategy, the policy included, is backtested as if it could trade at the closing
price it decides from. This flatters all arms alike, so comparisons between them are
fair, but absolute returns are optimistic. Momentum is the most sensitive: with a one-day
execution delay it returns $+214.7\%$, below buy-and-hold. The cloned networks are less
sensitive (Section~\ref{sec:clone}).
The 103 stocks are a hand-picked list of established companies that traded throughout
2016--2026, chosen partly with knowledge of their performance, and the S\&P~100 panel is a
survivor set in the same way. Absolute returns are therefore optimistic;
Section~\ref{sec:robust} bounds that but cannot remove it, and only a point-in-time
universe, which we do not have, would settle it. Every comparison in this paper uses the
same list, so comparisons between strategies are less affected, although a universe of
past winners may favour momentum. The cloning arm
is a capacity result whose teacher was chosen knowing it worked on this window. And the
test window is a strong bull market on TAIEX, the regime in which a passive bar is
hardest to beat. Because the portfolio is always fully invested, we cannot evaluate
whether a policy could add value by moving into cash; our conclusions concern stock
selection only.

These are properties of this setup. Whether they generalise beyond it is untested.

\section*{Code and data availability}

The analysis code will be made publicly available. Redistribution of the price series
depends on the vendor's terms.

\appendix
\label{app:start}

\section{Training and cloning configuration}
\label{app:config}

\paragraph{PPO training.} $1{,}200$ episodes on an 800-day sampling window, learning rate
$10^{-4}$, $\gamma=0.995$, GAE $\lambda=0.95$, clipping $0.2$, three epochs per update,
minibatch 128, entropy coefficient $10^{-3}$, and the router's load-balancing auxiliary
at $0.01$. The handoff runs of Section~\ref{sec:rq3} reuse all of these unchanged.

\paragraph{Imitation.} The loss is a cross-entropy from $\mathrm{softmax}$ of the stock
logits to the teacher's uniform-over-15 target, which is the only part of the action the
top-$k$ projection preserves. Two small extra terms: one (weight $0.1$) that keeps the
exposure output at full investment, which makes no difference to the portfolio because
the projection always invests fully, and the router's load-balancing penalty (weight
$0.01$), which keeps all four routed experts in use. Training runs three DAgger rounds
\citep{ross2011dagger} of 40 epochs each: plain behaviour cloning on the teacher's own
trajectory first, then two rounds in which the states the \emph{student} visits are
relabelled with what the teacher would have done there, which is what keeps the clone
from drifting off the distribution it was fitted on. Adam at $3\times10^{-4}$, batch 128,
validated every five epochs.

\section{Winner clustering, and why we hold it back}
\label{app:winners}

The models that beat the bar hold more alike than chance. Comparing each seed's daily
top-15 holding set, mean pairwise Jaccard similarity among the $29$ winners is $0.145$
against $0.107$ for random 29-subsets: $3.79$ shared names of 15 against $2.90$,
$z=+3.96$, $p=0.0016$ under a null that permutes which seeds carry the winner label with
the similarity matrix held fixed, so subset size enters observation and null identically.

Which bar defines a winner changes the figure, and against the reading we would prefer.
Run on the winner set defined by the \emph{uncharged} daily-rebalanced $1/N$ series the
set grows from 29 to 35 and the effect weakens to $z=+3.58$ ($p=0.0020$), because the six
models between that bar and buy-and-hold cluster at $z=-0.36$ ($p=0.56$) --- marginally
\emph{less} alike than chance, not more, so they dilute it. Against the charged bar the
set is 36; we did not re-run the permutation for it.

It survives the obvious artefacts. Measuring holdings on the validation window, which played no part
in defining the winners, gives $p=0.041$. Deleting the window's 20 best-performing stocks
moves $z$ to $+2.89$ ($p=0.011$). A jackknife loses significance in none of the 29
leave-one-out fits, dropping the two extreme seeds gives $p=0.0047$, and replacing the
Pearson profile correlation with a rank correlation gives $p=0.010$.

We nonetheless keep it out of the body, for two reasons that an untrained control makes
plain. Rebuilding the exact pre-training weights of all 100 seeds and rolling those
untrained networks over the same windows, the winners among them --- lucky rather than
skilled --- also hold more alike than a random pair. Winners are defined window by window
here, by each model's return over that window's uncharged daily-rebalanced $1/N$, so these
figures do not depend on which bar defines the $29$ above. Measured as holdings shared above a
random pair, out of 15, the untrained excess is $+0.46$, $+0.51$ and $+0.32$ on the 2022,
2023 and 2024 windows against a trained $+0.46$, $-0.38$ and $-0.55$: identical on the
first and of the opposite sign on the other two. Only on the test window does
training add overlap, $+0.69$ against an untrained $+0.15$ that does not reach
significance ($p=0.10$). The absolute level has a floor for the same reason: a random
pair of untrained networks already shares $2.40$ of its 15 daily holdings against $2.18$
for drawing 15 of 103 uniformly, while a random pair of trained models shares $2.90$.
A portion of winner concentration is therefore a property of lucky portfolios in this
architecture rather than of successful learning, and it is a statement about particular
regimes rather than about outperformance in general. Neither reading bears on whether the
policy has a selection advantage, which is what the body is about.

\section{Do similar portfolios earn similar returns?}

Across all $4{,}950$ seed pairs, the correlation between holding dissimilarity and the
absolute difference in test return is $-0.035$ (Pearson, $p=0.0152$) and $-0.038$
(Spearman, $p=0.0071$). Both are correlations of \emph{dis}similarity against the return gap, so the negative
sign says the opposite of the obvious reading: pairs holding \emph{less} alike earn more
alike. The magnitude says the relationship explains almost nothing either way --- what a
model holds is close to uninformative about how well it does --- and the sign is not stable
across windows, so we draw nothing from its direction.

\section{The momentum grid in full}

Table~\ref{tab:momgrid} prints the daily-rebalanced half of the grid, window by window
and lookback by lookback, and Table~\ref{tab:mom11} the other half, the same signal held
eleven days; together they are the $56$ cells per market the body counts. Holding longer cuts turnover from $0.194$ to $0.071$ per day and
therefore cost; it does not change the pattern.

\begin{table*}[t]\centering
\caption{Is the momentum baseline an artefact of one lookback on one window? Net excess
return over each window's own buy-and-hold, in percentage points, for top-15 momentum
rebalanced daily. Bold beats that bar. The grid is seven lookbacks $\times$ two holding
periods $\times$ four windows, 56 cells per market; the daily-rebalanced half is shown.}
\label{tab:momgrid}
\begin{tabular}{llrrrrrrrrr}
\toprule
market & window & B\&H & 5d & 10d & 20d & 40d & 60d & 120d & 250d & beats \\
\midrule
TAIEX & 2022 \emph{falling} & $-13.0\%$ & $-34$ & $-22$ & $-19$ & $-24$ & $-11$ & $-5$ & $-12$ & $0/7$ \\
TAIEX & 2023 \emph{bull} & $+56.4\%$ & $-27$ & $\mathbf{+11}$ & $\mathbf{+24}$ & $\mathbf{+30}$ & $\mathbf{+52}$ & $\mathbf{+20}$ & $\mathbf{+26}$ & $6/7$ \\
TAIEX & 2024 \emph{mild} & $+32.9\%$ & $-30$ & $-13$ & $-6$ & $\mathbf{+12}$ & $\mathbf{+24}$ & $\mathbf{+15}$ & $\mathbf{+22}$ & $4/7$ \\
TAIEX & test window & $+240.1\%$ & $\mathbf{+120}$ & $\mathbf{+162}$ & $\mathbf{+424}$ & $\mathbf{+110}$ & $\mathbf{+68}$ & $\mathbf{+7}$ & $\mathbf{+27}$ & $7/7$ \\
\midrule
S\&P 100 & 2022 \emph{falling} & $-7.9\%$ & $-23$ & $-20$ & $-14$ & $-4$ & $-6$ & $+0$ & $\mathbf{+1}$ & $1/7$ \\
S\&P 100 & 2023 \emph{bull} & $+15.4\%$ & $-12$ & $\mathbf{+3}$ & $-10$ & $\mathbf{+4}$ & $\mathbf{+6}$ & $\mathbf{+4}$ & $\mathbf{+6}$ & $5/7$ \\
S\&P 100 & 2024 \emph{mild} & $+16.7\%$ & $-30$ & $-23$ & $-19$ & $-12$ & $-3$ & $-5$ & $-4$ & $0/7$ \\
S\&P 100 & test window & $+25.9\%$ & $-29$ & $-23$ & $-10$ & $\mathbf{+8}$ & $-7$ & $-17$ & $\mathbf{+7}$ & $2/7$ \\
\bottomrule
\end{tabular}
\end{table*}

\begin{table*}[t]\centering
\caption{The eleven-day half of the momentum grid. Net excess over each window's own buy-and-hold, in percentage points. Bold beats that bar.}
\label{tab:mom11}
\begin{tabular}{llrrrrrrrr}
\toprule
market & window & 5d & 10d & 20d & 40d & 60d & 120d & 250d & beats \\
\midrule
TAIEX & 2022 \emph{falling} & $-16$ & $-12$ & $-15$ & $-20$ & $-7$ & $-3$ & $-8$ & $0/7$ \\
TAIEX & 2023 \emph{bull} & $-2$ & $\mathbf{+17}$ & $\mathbf{+36}$ & $\mathbf{+60}$ & $\mathbf{+42}$ & $\mathbf{+14}$ & $\mathbf{+22}$ & $6/7$ \\
TAIEX & 2024 \emph{mild} & $\mathbf{+0}$ & $\mathbf{+17}$ & $\mathbf{+7}$ & $\mathbf{+18}$ & $\mathbf{+22}$ & $\mathbf{+19}$ & $\mathbf{+20}$ & $7/7$ \\
TAIEX & test window & $\mathbf{+173}$ & $\mathbf{+307}$ & $\mathbf{+161}$ & $\mathbf{+59}$ & $-31$ & $\mathbf{+33}$ & $\mathbf{+88}$ & $6/7$ \\
\midrule
S\&P 100 & 2022 \emph{falling} & $-18$ & $-10$ & $-9$ & $-10$ & $-5$ & $\mathbf{+1}$ & $\mathbf{+7}$ & $2/7$ \\
S\&P 100 & 2023 \emph{bull} & $\mathbf{+2}$ & $+0$ & $+0$ & $\mathbf{+3}$ & $\mathbf{+3}$ & $\mathbf{+6}$ & $\mathbf{+11}$ & $5/7$ \\
S\&P 100 & 2024 \emph{mild} & $-11$ & $-9$ & $-1$ & $\mathbf{+2}$ & $-5$ & $\mathbf{+1}$ & $\mathbf{+10}$ & $3/7$ \\
S\&P 100 & test window & $\mathbf{+1}$ & $\mathbf{+18}$ & $\mathbf{+17}$ & $\mathbf{+24}$ & $\mathbf{+2}$ & $-7$ & $\mathbf{+19}$ & $6/7$ \\
\bottomrule
\end{tabular}
\end{table*}

\section{The interval for $P$}
\label{app:pint}

$P$ is a two-sample quantity, so both samples are resampled: every replicate draws 100
models with replacement and 200 matched random portfolios with replacement, and
recomputes $P$ over the $100\times200$ grid of that pair, $10{,}000$ times. Resampling one
side alone would understate the spread. Ties count as half a win, as they do in the
rank-sum. The result is $P=0.520$ gross with a $95\%$ interval of $[0.441, 0.600]$ and
$P=0.514$ net with $[0.435, 0.594]$. Both intervals contain $0.500$, and both are about
$\pm 0.08$ wide, which is the useful part: at 100 models and 200 draws this design can
exclude neither no advantage nor an advantage of the size a reader might care about. A
non-significant rank-sum alone would not have shown that.

\section{Cost sensitivity}
\label{app:costsens}

Table~\ref{tab:costsens} recosts every arm at the two realistic Taiwanese retail rates
of Section~\ref{sec:robust}.

\begin{table}[h]\centering\small
\caption{Net return at three per-leg rates: what the environment charges, and two
realistic Taiwanese retail rates. Recomputed from each arm's gross return and \emph{mean}
turnover, an approximation whose largest error against the exact figures is $1.33$ points
($0.148$ for momentum), which is why the $0.15\%$ column differs slightly from
Table~\ref{tab:costmatched}.}
\label{tab:costsens}
\begin{tabular}{lrrr}
\toprule
 & $0.15\%$ & $0.20\%$ & $0.29\%$ \\
\midrule
buy-and-hold & $+240.1\%$ & $+239.9\%$ & $+239.6\%$ \\
momentum, top 15 & $+308.2\%$ & $+294.8\%$ & $+271.7\%$ \\
random 15, matched & $+187.7\%$ & $+178.5\%$ & $+162.7\%$ \\
the 100 trained models & $+189.6\%$ & $+178.0\%$ & $+161.6\%$ \\
\quad beat B\&H & $29/100$ & $25/100$ & $22/100$ \\
the 32 cloned runs & $+341.1\%$ & $+320.1\%$ & $+287.3\%$ \\
\quad beat B\&H & $25/32$ & $24/32$ & $22/32$ \\
\bottomrule
\end{tabular}
\end{table}

\section{The tuning sweep}
\label{app:tuning}

The selection rule itself needs a noise floor, and has one. At the sweep stage, where
each arm was five seeds, the adopted arm cleared the base configuration by $13.4$ points
of validation excess --- a different quantity from the $+13.3\%$ against $+6.0\%$ above,
which are medians over 100 retrained seeds. Re-running the \emph{unchanged} configuration
as a do-nothing arm, also five seeds, reproduced $8.2$ of those $13.4$ points, and the two
are not distinguishable (rank-sum $p=0.69$; against five fresh base seeds instead the
adopted arm gives $p=0.55$, and with five against five the smallest attainable $p$ is
$0.008$, so neither comparison had the power to settle it). Most of what the rule selected
on is dispersion rather than improvement, which is why the 100-seed comparison, and not
the sweep, is what Section~\ref{sec:bar} relies on.

Raising the validation objective also consumes the one selection result the base
configuration had. Keeping the best quartile by validation score lifts the base
population's median excess over the uncharged daily-rebalanced $1/N$ to $+47\%$ and its
win rate against that series to $13/25$ ($p=0.036$); on the
tuned population the same rule carries no advantage at all (test median $-76.0\%$,
$9/25$, $p=0.76$), and the rank correlation between validation and test score falls from
$+0.247$ ($p=0.013$) to $+0.022$ ($p=0.83$). We report the effect and do not claim to
have characterised it.

\section{Determinism}
\label{app:determinism}

Training is not reproducible on this hardware at a fixed seed. Repeated runs of one
configuration and one seed, same device, span $-5.1\%$ to $+33.5\%$ on the validation
objective and $-83.1\%$ to $+306.7\%$ on test excess return --- a single-seed range
containing $72\%$ and $70\%$ of the range of the whole 100-seed population. Every claim
in this paper is therefore a population statement, and the cloning arm is counted per run
rather than per seed because one seed produced outcomes on both sides of the bar: its two
runs, same configuration and same device, returned $+183.1\%$ and $+280.8\%$.

\paragraph{It is not the seed.} Every random source the trainer draws from is seeded
before the model is built, and two processes at the same seed build bit-identical weights
while two different seeds do not. What the seed does not fix is the order in which the
arithmetic runs on the GPU, which the framework is free to vary between two runs of the
same code.

A mechanism the architecture makes available, which we state as a mechanism and have not
measured: the top-$k$ step thresholds on a strict inequality, so a difference in the last
bits of a score can change which names execute, and training is on-policy, so a changed
action changes the batch the next update is computed from. Over $1{,}200$ episodes there
is ample opportunity for a difference of that size to compound.

\paragraph{A single return is also not a sharp description of the policy.} Two cloned
runs sharing an initialisation hold $13.67$ of their 15 positions in common on an average
day, averaged over fourteen pairs, while their test returns differ by as much as $311.4$
percentage points. Nearly identical portfolios can therefore produce very different returns.

Within-seed reproducibility is recoverable, at a cost in speed, by fixing that order.
Doing so fixes replicates of one seed; it does nothing about the spread \emph{across}
seeds, which is the population this paper measures.

\section{The pre-registered pair-divergence replication}

Two cloned runs sharing an initialisation hold most of their positions in common while
their returns diverge widely, and an exploratory look at eight pairs suggested the two are
related: the rank correlation between a pair's return spread and its holding agreement was
$\rho=-0.905$ (exact $p=0.0046$), with the expert-partition version at $-0.833$
($p=0.0154$). Because that was found by looking at a table, we
pre-registered a replication on six fresh pairs with the predictions, the confidences and
the consequence of each outcome fixed in writing before the first of those runs reported.

All twelve runs have since reported and all five predictions held in sign. The
confirmatory arm does not reach significance on its own: $\rho=-0.771$ over six pairs
gives an exact two-sided $p$ of $0.1028$, and the partition version $0.1750$. A one-sided
test is defensible, since the direction was registered in advance, and would give
$0.0514$; we did not register one and do not take it. What replicated is the sign, on
fresh data, with the magnitude attenuating as predicted. Pooling both arms gives
$\rho=-0.886$ over fourteen pairs, but eight of those rows generated the hypothesis, so
that figure is an estimate and not a test.

\section{Injected delistings}
\label{app:delist}

Deleting the best full-sample performers, which is the ex-post information survivorship
smuggles in, \emph{widens} momentum's edge over uncharged daily-rebalanced $1/N$ from
$+89.3$ to $+102.9$, $+111.8$ and $+126.8$ points at 5, 10 and 20 names (a separate
simulation, measured against the uncharged series rather than buy-and-hold). But that tests only whether survivorship flatters
momentum; it cannot put back what a survivor panel removes, which is names that fell and
stopped trading. So we inject them. For each of 200 draws per cell we delist $k$ names on
uniformly chosen days with a terminal return of $-X\%$, make them untradeable afterwards,
and re-run three arms over the modified universe: buy-and-hold, momentum re-selected
daily on the modified prices, and the 100 policies' recorded daily holdings replayed.
Sweeping $k\in\{5,10,20\}$ and $X\in\{30,60,100\}$, momentum's margin over buy-and-hold
stays between $+61.0$ and $+90.3$ points against $+68.3$ on the unmodified panel, which
this harness reproduces exactly, so levels as well as changes between panels can be read.
Momentum beats the bar in $94$ to $100\%$ of draws in every cell, and the policy
population wins $33$ to $43$ times in $100$ --- never a majority. The bar itself falls far
more than the margin does, from $+240.1\%$ to as little as $+174.9\%$: a universe
containing losers hurts an equal-weighted passive portfolio holding all 103 names more
than it hurts a concentrated rule that can rotate out of them. At the extreme cell, total
wipeout of 20 names, the margin's fifth percentile touches $-1.6$ points, so an individual
draw can erase it.

Two limits on that simulation. The policy arm replays each policy's
recorded holdings, equal-weighted, an approximation that reads about 14 points high, so
only the comparison between universes should be read from its level. And the policies cannot react to a delisting
they never saw while momentum can. That favours momentum, so its surviving margin is, if
anything, optimistic, and the policy population's shortfall may be overstated.

\section{Restricting to the names on the list before the test window}
\label{app:pretest}

The 103 names are a hand-maintained list, grown in dated batches, and $38$ of them
were added to it in batches dated 2026-04 to 2026-06 --- inside the test window, which
runs from January 2025. The criterion here is that list and its batch dates, not a
listing date from the exchange: every one of the 103 has a price on every day of the
sample (Section~\ref{sec:robust}), so what the late batches introduce is look-ahead in
the choice of universe rather than in any price. The other $65$ names were on the list
before the window opened. Re-running
the comparisons on those alone changes none of the paper's conclusions
(Table~\ref{tab:pretest}): momentum still beats the bar, its margin narrowing from
$+68.3$ to $+57.6$ points, and the trained population is still nowhere near a majority.
Six clones retrained from scratch on the restricted panel beat its buy-and-hold in
\textbf{five of six} at the validation-selected checkpoint (median $+292.7\%$) and in
\textbf{six of six} at the end of cloning (median $+398.8\%$), against $25$ of $32$ and
$31$ of $32$ on the full panel. On the restricted panel each of the twelve checkpoints
earns $1.5$ to $7.0$ times its own turnover cost, where the teacher manages $3.5$ --- the
same teacher earns $4.1$ times its cost on the full panel, so the two ratios are not
comparable across panels.

\begin{table}[h]\centering\small
\caption{Every arm on the full panel and on the $65$ names that were on the list before
the test window, net over the same 345 scored steps. Margins quoted in the text are
differences of unrounded values, so they can differ by $0.1$ from a difference of the
figures printed here. The matched-random row uses the same $200$ seeds and the same redraw
probability as Table~\ref{tab:costmatched}, so on the full panel it reproduces that row;
on the restricted panel those seeds draw from the $65$ names instead. The policy
row replays each model's recorded holdings, equal-weighted; that approximation reads about
14 points high and holds a mean of $8.69$ of 15 names once restricted, so only its change
between panels should be read.}
\label{tab:pretest}
\begin{tabular}{lrr}
\toprule
 & all 103 & on the list before \\
\midrule
buy-and-hold          & $+240.1\%$ & $+243.5\%$ \\
momentum, top 15      & $+308.4\%$ & $+301.2\%$ \\
random 15, matched    & $+189.1\%$ & $+198.1\%$ \\
policy replay         & $+204.1\%$ & $+207.3\%$ \\
\quad beat B\&H       & $36/100$   & $36/100$ \\
\bottomrule
\end{tabular}
\end{table}

\section{The two handoff screens in full}
\label{app:screens}

Both screens of Section~\ref{sec:screens} ran the same eight cloned checkpoints for 600
episodes. \emph{Agreement} is the mean
teacher overlap over the deterministic evaluations at episodes 550, 575 and 600, $R$ the
mean objective over the same three, $R_{\text{clone}}$ that run's episode-1 value, and
$R_{\text{samp}}$ the mean sampled episode reward over the last fifty episodes.
Table~\ref{tab:screenruns} gives all twenty-four runs.

\begin{table}[h]\centering\footnotesize
\caption{Per-run outcomes. No run in any arm satisfies the pre-registered rule
(agreement $\geq 7$ and $R \geq R_{\text{clone}}-2$).}
\label{tab:screenruns}
\begin{tabular}{llrrrr}
\toprule
arm & run & agree. & $R$ & $\Delta R$ & $R_{\text{samp}}$ \\
\midrule
base & 1 & $2.43$ & $-8.79$ & $-9.27$ & $-11.90$ \\
base & 2 & $4.86$ & $-9.15$ & $-2.76$ & $-15.24$ \\
base & 3 & $2.28$ & $-11.61$ & $-17.71$ & $-10.60$ \\
base & 4 & $3.22$ & $-10.28$ & $-7.81$ & $-12.40$ \\
base & 5 & $5.96$ & $-14.12$ & $-15.42$ & $-17.08$ \\
base & 6 & $3.98$ & $-12.12$ & $-26.68$ & $-10.34$ \\
base & 7 & $1.15$ & $-6.02$ & $-8.63$ & $-10.04$ \\
base & 8 & $1.94$ & $-6.71$ & $-1.74$ & $-12.64$ \\
\midrule
B1 & 1 & $4.43$ & $-12.38$ & $-12.87$ & $-10.88$ \\
B1 & 2 & $1.02$ & $+1.04$ & $+7.43$ & $-4.42$ \\
B1 & 3 & $1.36$ & $-21.15$ & $-27.25$ & $-12.24$ \\
B1 & 4 & $0.54$ & $-22.61$ & $-20.14$ & $-15.81$ \\
B1 & 5 & $2.64$ & $-17.39$ & $-18.69$ & $-12.16$ \\
B1 & 6 & $2.39$ & $-10.72$ & $-25.28$ & $-10.93$ \\
B1 & 7 & $5.68$ & $-15.14$ & $-17.75$ & $-10.59$ \\
B1 & 8 & $3.58$ & $-15.20$ & $-10.23$ & $-6.69$ \\
\midrule
A & 1 & $0.52$ & $-12.29$ & $-12.77$ & $-10.94$ \\
A & 2 & $0.14$ & $+12.42$ & $+18.82$ & $+1.74$ \\
A & 3 & $0.22$ & $+20.58$ & $+14.47$ & $+9.94$ \\
A & 4 & $0.92$ & $+10.66$ & $+13.13$ & $+11.17$ \\
A & 5 & $0.42$ & $-20.00$ & $-21.29$ & $-9.64$ \\
A & 6 & $0.22$ & $+21.42$ & $+6.86$ & $+9.16$ \\
A & 7 & $0.08$ & $+20.92$ & $+18.31$ & $+9.34$ \\
A & 8 & $0.52$ & $-12.96$ & $-7.99$ & $-13.52$ \\
\bottomrule
\end{tabular}
\end{table}

Three readouts did not fit the body. Screen A's sampled turnover over the last fifty
episodes is $0.23$ of the portfolio a day against the baseline's $0.54$, while its exploration spread rises
to $\sigma=0.637$ from the inherited $0.368$ --- the Plackett--Luce policy loss has no
gradient in that spread at all, so only the entropy bonus acts on it. Screen A's
importance ratio is also far better behaved, with $38\times$ less dispersion --- spread
across the sampled actions in a batch --- than the ratio over the full
$104$-dimensional action, over the first ten updates, lower in 8 of 8 runs. Test returns are too noisy to adjudicate any of this: screen A's median is
$+430\%$ at the validation-selected checkpoint and $+168\%$ at the unselected final one,
so the checkpoint rule alone moves it by $262$ points. The two rules disagree in the
baseline too --- across its eight runs the unselected checkpoint wins four, ties two and
loses two, with medians $+272\%$ against $+178\%$.

One prediction cannot be scored. The pre-registration asked whether screen A clips more
often than the baseline over the first ten updates, reasoning that a likelihood over 15
names moves faster per update than one over the raw action's 104 coordinates. That comparison is not available for
screen A, so we report the failure rather than substituting a different test. What can be
compared points the other way: the baseline's $4.8\%$ clip fraction is largely noise
across the $89$ coordinates
the projection makes irrelevant, and scoring only the executed names removes it.

\section{Glossary and notation}
\label{app:glossary}

Table~\ref{tab:glossary} gathers the terms this paper uses in a technical sense. Each is
also defined in the body where it first appears; the table is for reading out of order.

\begin{table*}[t]\centering\footnotesize
\caption{Glossary and notation.}
\label{tab:glossary}
\begin{tabular}{p{0.21\textwidth}p{0.62\textwidth}p{0.09\textwidth}}
\toprule
term & meaning & where \\
\midrule
basis point & One hundredth of a percentage point; the unit for per-day edges and costs. & \S\ref{sec:setup} \\
turnover $T_t$ & The sum of $|w_{i,t}-w_{i,t-1}|$ over the 103 stocks and cash, measured target to target, so drift between steps is free. Cash is zero throughout, so only the stocks contribute; this is not the action's $104$ coordinates. & \S\ref{sec:setup} \\
gross vs net & Net is the return after the trading charge; gross refunds that charge along the identical holding path, isolating selection from trading. & \S\ref{sec:bar} \\
buy-and-hold & $1/N$ bought on the first test day and held to the last, weights drifting with returns; nothing to charge after day one. & \S\ref{sec:setup} \\
daily-rebalanced $1/N$ & Equal weight restored every day; the series the environment ships, quoted charged or uncharged as stated. & \S\ref{sec:setup} \\
comparison series & The uncharged daily-rebalanced $1/N$ return $r^{1/N}_t$ that the reward subtracts each day; a baseline the policy cannot move, so it scores $R=0$. & \S\ref{sec:setup} \\
teacher overlap & How many of the 15 stocks a model holds are also among the momentum teacher's 15 picks, averaged over days; $15$ is identical holdings and two random 15-name sets share about $15\times15/103\approx2.18$. & \S\ref{sec:clone} \\
momentum (the teacher) & Hold the 15 names with the highest trailing 60-day return, rebalanced daily and charged like every other arm. & \S\ref{sec:bar} \\
selection edge & The extra return per day the names held earn over the universe average, before its own trading cost. & \S\ref{sec:bar} \\
information coefficient & The cross-sectional rank correlation between the weights the policy assigns and the returns that follow. & \S\ref{sec:bar} \\
probability of superiority $P$ & The chance that a randomly chosen member of one population beats a randomly chosen member of another; an even chance is no advantage. & \S\ref{sec:bar} \\
top-$k$ projection & The step applied to every action before the environment executes it: keep the 15 largest stock weights, renormalise them to sum to one, set cash to zero. & \S\ref{sec:setup} \\
exposure head & The head that decides how much of the portfolio to invest; the projection discards it, so it cannot move the executed portfolio. & \S\ref{sec:rq4} \\
critic / value head & The half of the actor-critic that predicts future reward and supplies PPO's baseline; the imitation loss never touches it. & \S\ref{sec:setup}, \S\ref{sec:rq3} \\
exploration noise & The spread of the Normal the policy samples its actions from, which is itself a learned parameter; cloning leaves it at its default. & \S\ref{sec:rq4} \\
the training objective $R$ & What PPO ascends: $R_t = 15[\log(1+r^{\text{net}}_t)-\log(1+r^{1/N}_t)]$, summed over a window's scored steps, so $R=0$ is the comparison series itself. & \S\ref{sec:setup}, \S\ref{sec:rq3} \\
importance ratio & The new policy's probability of the stored action over the old policy's; the quantity PPO clips. Its \emph{dispersion} is the spread of that ratio across a batch. & \S\ref{sec:screens} \\
Plackett--Luce likelihood & The probability of a ranked list drawn one item at a time without replacement; screen A scores the executed top-15 with it. & \S\ref{sec:screens} \\
behaviour cloning & Supervised fitting of the policy to a teacher's actions along the teacher's own trajectory. & \S\ref{sec:clone}, App.~\ref{app:config} \\
DAgger relabelling & Further rounds in which the states the \emph{student} visits are labelled with what the teacher would have done there. & App.~\ref{app:config} \\
validation-selected checkpoint & Weights chosen on a trailing mean of the last three validation scores. & \S\ref{sec:clone} \\
end-of-training checkpoint & The weights at the end of training, selected on nothing. & \S\ref{sec:clone} \\
survivorship & Every name in a panel has a price on every day, so names that failed and stopped trading are absent and every absolute return is inflated. & \S\ref{sec:robust} \\
\bottomrule
\end{tabular}
\end{table*}

\bibliographystyle{ACM-Reference-Format}
\bibliography{refs}

\end{document}